\documentclass[runningheads]{llncs}

\usepackage[T1]{fontenc}
\usepackage{graphicx}
\usepackage{multirow}
\usepackage[dvipsnames]{xcolor}
\usepackage{etoolbox}
\usepackage{amsmath}
\usepackage{latexsym,amssymb}
\usepackage{upgreek}
\usepackage[noend]{algpseudocode}
\usepackage{array}
\usepackage{adjustbox}
\usepackage{soul}
\usepackage{url}
\usepackage[utf8]{inputenc}
\usepackage{booktabs}
\usepackage{empheq}
\usepackage{fancybox}
\usepackage{subcaption}
\usepackage[english]{babel}
\usepackage{xspace}
\usepackage{setspace}
\usepackage{stmaryrd}
\usepackage{enumitem}
\usepackage{ifthen}
\usepackage{verbatim}
\usepackage{algorithm}
\usepackage{algorithmicx}
\let\llncssubparagraph\subparagraph
\let\subparagraph\paragraph
\usepackage{titlesec}
\let\subparagraph\llncssubparagraph

\usepackage{amsthm}
\usepackage{hyperref}
\usepackage[capitalise,nameinlink]{cleveref}
\usepackage{xfrac}
\usepackage{nicefrac}       
\usepackage{makecell}
\usepackage{bbding}

\usepackage{tikz}
\usepackage{forest}

\usetikzlibrary{arrows,decorations.pathmorphing,decorations.footprints,fadings,calc,trees,mindmap,shadows,decorations.text,patterns,positioning,shapes,matrix,fit}
\usetikzlibrary{arrows,shadows,backgrounds}
\usetikzlibrary{arrows.meta}
\usetikzlibrary{positioning,fit}
\usetikzlibrary{automata}
\usetikzlibrary{matrix}
\usetikzlibrary{shapes.symbols,shapes.misc,shapes.arrows,shapes.geometric}
\usetikzlibrary{matrix,chains,scopes,decorations.pathmorphing}

\newboolean{nonanon}               
\setboolean{nonanon}{true}         %
\newboolean{mkcompact}             
\setboolean{mkcompact}{true}       %
\newboolean{squeezeit}             
\setboolean{squeezeit}{true}       %
\hypersetup{
  colorlinks = true,
  linkcolor = darkblue,
  urlcolor  = darkblue,
  citecolor = darkblue,
  anchorcolor = darkblue
}

\crefname{prop}{Proposition}{Propositions}
\Crefname{prop}{Proposition}{Propositions}
\crefname{defn}{Definition}{Definitions}
\Crefname{defn}{Definition}{Definitions}
\crefname{cor}{Corollary}{Corollaries}
\Crefname{cor}{Corollary}{Corollaries}
\crefname{exmpl}{Example}{Examples}
\Crefname{exmpl}{Example}{Examples}
\Crefname{theorem}{Theorem}{Theorem}
\crefname{page}{page}{page}

\newtheoremstyle{nutstyle}
{3.5pt} 
{3.5pt} 
{} 
{} 
{\bfseries} 
{.} 
{.5em} 
{} 

\theoremstyle{nutstyle}
\newtheorem{prop}{Proposition}

\newtheoremstyle{nuxstyle}
{3.5pt} 
{3.5pt} 
{} 
{} 
{\itshape} 
{.} 
{.5em} 
{} 

\theoremstyle{nuxstyle}
\newtheorem{exmpl}{Example}

\newcommand{\todoF}[2]{}

\newcommand{\fml}[1]{{\mathcal{#1}}}

\newcommand{\tn}[1]{\textnormal{#1}}
\newcommand{\mbf}[1]{\ensuremath\mathbf{#1}}
\newcommand{\msf}[1]{\ensuremath\mathsf{#1}}
\newcommand{\mbb}[1]{\ensuremath\mathbb{#1}}

\newcommand{\tbf}[1]{\textbf{#1}}

\newcommand{\stwop}{\Upsigma_2^\tn{P}}
\newcommand{\ptwop}{\Uppi_2^\tn{P}}

\definecolor{darkslategray}{rgb}{0.18, 0.31, 0.31} 
\definecolor{platinum}{rgb}{0.9, 0.89, 0.89} 
\definecolor{gray}{rgb}{.4,.4,.4}
\definecolor{midgrey}{rgb}{0.5,0.5,0.5}
\definecolor{middarkgrey}{rgb}{0.35,0.35,0.35}
\definecolor{darkgrey}{rgb}{0.3,0.3,0.3}
\definecolor{darkred}{rgb}{0.7,0.1,0.1}
\definecolor{midblue}{rgb}{0.2,0.2,0.7}
\definecolor{darkblue}{rgb}{0.1,0.1,0.5}
\definecolor{defseagreen}{cmyk}{0.69,0,0.50,0}

\definecolor{medred}{rgb}{0.5,0.1,0.1}
\definecolor{midred}{rgb}{0.7,0.2,0.2}
\definecolor{vdarkred}{rgb}{0.4,0.1,0.1}
\definecolor{darkslategray}{rgb}{0.18, 0.31, 0.31} 
\definecolor{platinum}{rgb}{0.9, 0.89, 0.89} 
\definecolor{gray}{rgb}{.4,.4,.4}
\definecolor{darkred}{rgb}{0.7,0.1,0.1}
\definecolor{darkblue}{rgb}{0.1,0.1,0.5}
\definecolor{darkgreen}{rgb}{0.1,0.5,0.1}
\definecolor{purple3}{RGB}{125,38,205}          

\definecolor{tyellow1}{HTML}{FCE94F}
\definecolor{tyellow2}{HTML}{EDD400}
\definecolor{tyellow3}{HTML}{C4A000}
\definecolor{torange1}{HTML}{FCAF3E}
\definecolor{torange2}{HTML}{F57900}
\definecolor{torange3}{HTML}{C35C00}
\definecolor{tbrown1}{HTML}{E9B96E}
\definecolor{tbrown2}{HTML}{C17D11}
\definecolor{tbrown3}{HTML}{8F5902}
\definecolor{tgreen1}{HTML}{8AE234}
\definecolor{tgreen2}{HTML}{73D216}
\definecolor{tgreen3}{HTML}{4E9A06}
\definecolor{tblue1}{HTML}{729FCF}
\definecolor{tblue2}{HTML}{3465A4}
\definecolor{tblue3}{HTML}{204A87}
\definecolor{tpurple1}{HTML}{AD7FA8}
\definecolor{tpurple2}{HTML}{75507B}
\definecolor{tpurple3}{HTML}{5C3566}
\definecolor{tred1}{HTML}{EF2929}
\definecolor{tred2}{HTML}{CC0000}
\definecolor{tred3}{HTML}{A40000}
\definecolor{tlgray1}{HTML}{EEEEEC}
\definecolor{tlgray2}{HTML}{D3D7CF}
\definecolor{tlgray3}{HTML}{BABDB6}
\definecolor{tdgray1}{HTML}{888A85}
\definecolor{tdgray2}{HTML}{555753}
\definecolor{tdgray3}{HTML}{2E3436}

\newcommand{\dghlight}[1]{{\color[RGB]{0,120,0}#1}}

\newcommand{\jnoteF}[1]{}
\newcommand{\anoteF}[1]{}

\newcounter{tableeqn}[table]

\DeclareMathOperator*{\limply}{\rightarrow}

\newcommand{\waxp}{\ensuremath\mathsf{WAXp}}
\newcommand{\wcxp}{\ensuremath\mathsf{WCXp}}
\newcommand{\axp}{\ensuremath\mathsf{AXp}}
\newcommand{\cxp}{\ensuremath\mathsf{CXp}}
\newcommand{\aex}{\ensuremath\mathsf{AEx}}

\newcommand{\relevant}{\small{\mathsf{Relevant}}}
\newcommand{\irrelevant}{\small{\mathsf{Irrelevant}}}
\newcommand{\yesmark}{{\small\Checkmark}}
\newcommand{\nomark}{{\small\XSolidBrush}}

\newcommand{\exv}{\ensuremath\mathbf{E}}

\newcommand{\prob}{\ensuremath\mathbf{P}}

\newcommand{\cf}{\ensuremath\upsilon} 
\newcommand{\cfn}[1]{\ensuremath\upsilon_{#1}} 

\newcommand{\svn}[1]{\msf{Sc}_{#1}}

\newcommand{\similar}{\ensuremath\sigma}
\newcommand{\tsimilar}{\ensuremath\mathsf{T}\sigma}

\newcommand{\rows}{\ensuremath\msf{rows}}

\newcommand{\mailtodomain}[1]{\href{mailto:#1@ciencias.ulisboa.pt}{\texttt{\nolinkurl{#1}}}}

\DeclareMathOperator*{\sv}{\msf{Sc}}

\titleformat{\paragraph}[runin]
{\normalfont\bfseries}{}{0pt}{}

\newcolumntype{L}[1]{>{\raggedright\let\newline\\\arraybackslash\hspace{0pt}}m{#1}}
\newcolumntype{C}[1]{>{\centering\let\newline\\\arraybackslash\hspace{0pt}}m{#1}}
\newcolumntype{R}[1]{>{\raggedleft\let\newline\\\arraybackslash\hspace{0pt}}m{#1}}

\definecolor{dkgreen}{rgb}{0,0.6,0}
\definecolor{ltblue}{rgb}{0,0.4,0.4}
\definecolor{dkviolet}{rgb}{0.3,0,0.5}

\tikzset{
  0 my edge/.style={densely dashed, my edge, draw=midblue},
  my edge/.style={-{Stealth[]}, draw=midblue},
}

\makeatletter
\renewcommand{\paragraph}{%
  \@startsection{paragraph}{4}%
  {\z@}{1.75ex \@plus 1ex \@minus .2ex}{-1em}%
  {\normalfont\normalsize\bfseries}%
}
\makeatother

\begin{document}
%

\title{Logic-Based Explainability:\\Past, Present \& Future}
\ifthenelse{\boolean{nonanon}}{
  \titlerunning{Logic-Based XAI}
}{
  \titlerunning{}
}
%

\ifthenelse{\boolean{nonanon}}{
  \author{%
    Joao Marques-Silva\inst{1}\orcidID{0000-0002-6632-3086}
  }
  \authorrunning{J. Marques-Silva}
  %
  \institute{%
    ICREA, University of Lleida, Spain
    \email{jpms@icrea.cat}
  }
}{
  \author{\tbf{Paper \# NNN}}
  \authorrunning{~~}
  \institute{}
}

\maketitle              

\begin{abstract}
  In recent years, the impact of machine learning (ML) and artificial
  intelligence (AI) in society has been absolutely remarkable. This
  impact is expected to continue in the foreseeable future. However,
  the adoption of AI/ML is also a cause of grave concern. The
  operation of the most advances AI/ML models is often beyond the
  grasp of human decision makers. As a result, decisions that impact
  humans may not be understood and may lack rigorous validation. 
  Explainable AI (XAI) is concerned with providing human
  decision-makers with understandable explanations for the predictions
  made by ML models. As a result, XAI is a cornerstone of trustworthy
  AI.
  Despite its strategic importance, most work on XAI lacks rigor, and
  so its use in high-risk or safety-critical domains serves to foster
  distrust instead of contributing to build much-needed trust.
  Logic-based XAI has recently emerged as a rigorous alternative to
  those other non-rigorous methods of XAI.
  This paper provides a technical survey of logic-based XAI, its
  origins, the current topics of research, and emerging future topics
  of research. The paper also highlights the many myths that pervade
  non-rigorous approaches for XAI.
  %

\end{abstract}

\section{Introduction} \label{sec:intro}

\jnoteF{Remarkable progress in ML\&AI, and astonishing pace of progress
  in recent years.\\
  But, the ML models of choice are without exception inscrutable.
  This means predictions made by ML models cannot de fathomed, often
  not even by domain experts. Of course, such predictions could be
  plain wrong, e.g. because of bugs in the data, or in the training
  algorithm.\\
  There is also regulation, already in place and upcoming, which puts
  demands on the uses of ML models, but also the needs for their
  operation to be understood~\cite{hlegai19,wheo23,euaict24}.\\
  Explainability aims to bridge the gap between highly complex ML
  models and the need to explain models.
}

The last decade witnessed remarkable advances in Machine Learning
(ML) and Artificial Intelligence (AI). According to different
estimates, the AI/ML market is forecast to  increase by more than a
ten fold until the end of the decade.%
\footnote{
E.g.\ 
\url{https://tinyurl.com/3xswxma4},
\url{https://tinyurl.com/28nse7ms},
\url{https://tinyurl.com/bdctfmhy},
\url{https://tinyurl.com/msszbk4m},
\url{https://tinyurl.com/4hem53z7},
\url{https://tinyurl.com/mvucuc4b}.
}
Nevertheless, trust represents the major obstacle to widespread
deployment of AI/ML models. The best performing AI/ML models operate
as black-boxes, and most often their results cannot be fathomed by
human decision-makers. However, in many uses of ML, it is critical
to be able to explain and/or debug decisions, such that these can be
reasoned about. More importantly, in high-risk and safety-critical
domains, the rigor of explanations is paramount.
The importance of trust in the operation of AI/ML models has led to
expert recommendations~\cite{hlegai19} and, more recently, to
legislation~\cite{wheo23,euaict24} aiming at regulating the validation 
and operation of models of AI/ML. The urgency of measures is further
underscored by joint declarations of different countries at recent AI
summits.%
\footnote{
E.g.\ 
\url{https://tinyurl.com/bddr6yf5},
\url{https://tinyurl.com/44d5ywxt}.
}

Explainable AI (XAI) is at the core of efforts to deliver trustworthy
AI~\cite{hlegai19,seshia-cacm22}.
Accordingly, there have been massive efforts towards devising methods
of XAI over the last few years.
Nevertheless, most of these efforts do not offer guarantees of rigor,
and so their use in high-risk and/or safety-critical domains is at
best problematic.
As past experience with the use of formal methods amply confirms,
trustable AI/ML requires rigorous approaches to explainability, including
rigorous definitions, rigorous computation of explanations, but also
the certification of computed results.
The lack of rigor of most existing methods of XAI motivated the
emergence of the field of formal XAI since
2018/2019~\cite{darwiche-ijcai18,inms-aaai19}.
In contrast with most existing approaches to XAI, formal XAI is based
on rigorous logic-based definitions of explanations, being
model-precise.

The main goal of this paper is to provide an update on the ongoing
progress in formal XAI, extending other recent
overviews~\cite{msi-aaai22,ms-rw22,msi-frai23}. 
Furthermore, the paper also provides a glimpse of the ongoing efforts
to unify the two main approaches for XAI: explainability by feature
attribution and explainability by feature selection.

The paper is organized as follows.
\cref{sec:prelim} introduces the definitions and notation used in the
remainder of the paper.
\cref{sec:past} overviews the recent past in formal explainability,
covering the core foundations.
\cref{sec:present} overviews ongoing topics of research related with
formal explainability.
\cref{sec:myths} overviews how formal explainability has enabled
disporving several misconceptions of non-rigorous approaches to XAI.
\cref{sec:glimpse} briefly overviews near future lines of research.
Finally, \cref{sec:conc} concludes the paper.

\section{Preliminaries} \label{sec:prelim}


\paragraph{Logic foundations.}
%
The paper assumed basic knowledge of propositional and first-order
logic. The notation used is standard in recent
references~\cite{mc-handbook18,sat-handbook21}.
This includes basic knowledge of the decision problem for
propositional logic, i.e.\ the well-known Satisfiability (SAT)
problem~\cite{sat-handbook21}, but also of (decidable) fragments of
first-order logic, which are often referred to as Satisfiability
Modulo Theories (SMT)~\cite{mc-handbook18,sat-handbook21}. In
addition, basic knowledge of mixed integer linear programming (MILP)
models and reasoners is assumed.

Logic-based explainability builds on logic in the following
ways.
First, logic encodings are used for reasoning about ML models.
Second, automated reasoners (e.g.\ SAT, SMT, MILP reasoners) are used
as oracles for solving a multitude of computational problems.
Third, logic is used for formalizing concepts in explainability.
Furthermore, more specific topics, including prime
implicants~\cite{crama-bk11}, minimal unsatisfiable subsets
(MUSes)~\cite{sat-handbook21} and minimal correction
subsets~\cite{sat-handbook21} are also used extensively in logic-based
XAI.%
\footnote{%
Furthermore, it is important to deconstruct a persisting misconception
in some AI\&ML bibliography. Indeed, there is a widespread belief that
theoretical computational intractability (e.g. NP-hardness) equates
with practical computational intractability. Although this 
observation is formally correct \emph{in the worst-case}, it is often
\emph{not} the case in many practical settings. A well-known example
is the Simplex algorithm for LP, which is worst-case exponential in
the worst-case, but that is hardly ever the case in
practice~\cite{minty-ineq72,borgwardt-zor82}. 
Other inevitable examples include SAT, SMT and MILP
reasoners~\cite{sat-handbook21,mc-handbook18}. These reasoners have
been the subject of remarkable progress over the last two
decades~\cite{mc-handbook18,sat-handbook21}. Some of these advances in
automated reasoners resulted from the author's work on SAT solvers in
the mid and late
1990s~\cite{mss-iccad96,mss-tcomp99,msm-hbk18,mslm-hbk21}.}

\jnoteF{The standard definitions apply. But we briefly overview
  less-known topics, including primes, MUSes, MCSes and their
  enumeration.}

\jnoteF{Clarify the misconception of practical \emph{intractability}.}

\jnoteF{Progress in automated reasoning.}


\jnoteF{A persisting misconception in some AI\&ML bibliography is that
  theoretical intractability (e.g. NP-hardness) equates with practical
  computational intractability. Although this observation \emph{in the
  worst-case}, it is often not the case in most practical settings.
  A well-known example is the Simplex algorithm for LP. Other
  inevitable examples include the decision problem for propositional
  logic (i.e.\ the SAT problem) and mixed integer linear programming.}

\paragraph{Classification \& regression problems.}
%
Let $\fml{F}=\{1,\ldots,m\}$ denote a set of features.
Each feature $i\in\fml{F}$ takes values from a domain $\mbb{D}_i$.
Domains can be categorical or ordinal. If ordinal, domains can be
discrete or real-valued.
Feature space is defined by
$\mbb{F}=\mbb{D}_1\times\mbb{D}_2\times\ldots\times\mbb{D}_m$. 
Throughout the paper domains are assumed to be discrete-valued.%
\footnote{
The results in the paper can be generalized to real-valued features.
For real-valued features, the only changes involve the definition of
expected value and probability, respectively in~\eqref{eq:evdef}
and~\eqref{eq:probdef}.}
The notation $\mbf{x}=(x_1,\ldots,x_m)$ denotes an arbitrary point in 
feature space, where each $x_i$ is a variable taking values from
$\mbb{D}_i$. Moreover, the notation $\mbf{v}=(v_1,\ldots,v_m)$
represents a specific point in feature space, where each $v_i$ is a
constant representing one concrete value from $\mbb{D}_i$.
A classifier maps each point in feature space to a class taken from
$\fml{K}=\{c_1,c_2,\ldots,c_K\}$. Classes can also be categorical or
ordinal. However, and unless otherwise stated, classes are assumed to
be ordinal.
In the case of regression, each point in feature space is mapped to an
ordinal value taken from a set of values $\mbb{V}$, e.g.\ $\mbb{V}$
could denote $\mbb{Z}$ or $\mbb{R}$.
Therefore, a classifier $\fml{M}_{C}$ is characterized by a
non-constant \emph{classification function} $\kappa$ that maps feature
space $\mbb{F}$ into the set of classes $\fml{K}$,
i.e.\ $\kappa:\mbb{F}\to\fml{K}$.
A regression model $\fml{M}_R$ is characterized by a non-constant
\emph{regression function} $\rho$ that maps feature space $\mbb{F}$
into the set elements from $\mbb{V}$, i.e.\ $\rho:\mbb{F}\to\mbb{V}$. 
A classifier model $\fml{M}_{C}$ is represented by a tupple
$(\fml{F},\mbb{F},\fml{K},\kappa)$, whereas a regression model
$\fml{M}_{R}$ is represented by a tupple
$(\fml{F},\mbb{F},\mbb{V},\rho)$.
When viable, we will represent an ML model $\fml{M}$ by a tupple
$(\fml{F},\mbb{F},\mbb{T},\tau)$, with $\tau:\mbb{F}\to\mbb{T}$,
without specifying whether whether $\fml{M}$ denotes a classification
or a regression model.
A \emph{sample} (or instance) denotes a pair $(\mbf{v},q)$, where
$\mbf{v}\in\mbb{F}$ and either $q\in\fml{K}$, with
$q=\kappa(\mbf{v})$, or $q\in\mbb{V}$, with $q=\rho(\mbf{v})$.

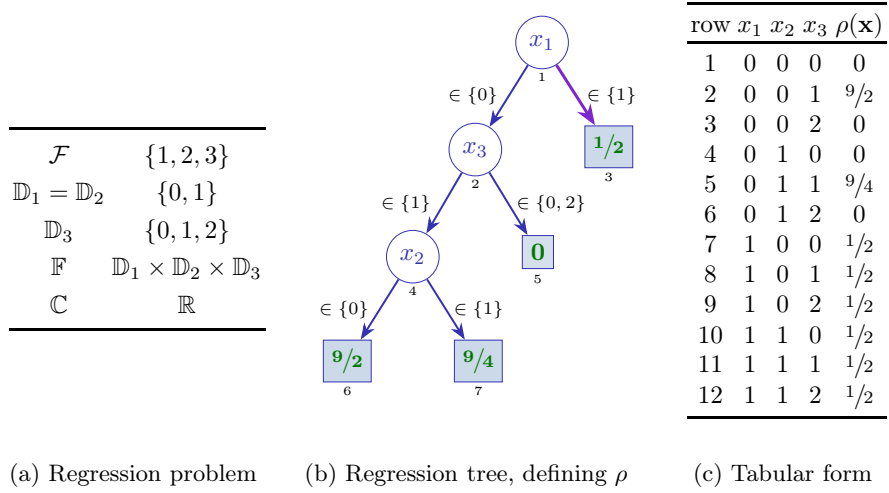
\begin{figure*}[t]
  \begin{subfigure}[b]{0.3125\textwidth}
    \begin{center}
      \scalebox{0.95}{
        \renewcommand{\arraystretch}{1.25}
        \begin{tabular}{cc} \toprule
          $\fml{F}$ & $\{1,2,3\}$ \\
          $\mbb{D}_1=\mbb{D}_2$ & $\{0,1\}$ \\
          $\mbb{D}_3$ & $\{0,1,2\}$ \\
          $\mbb{F}$ & $\mbb{D}_1\times\mbb{D}_2\times\mbb{D}_3$ \\
          $\mbb{C}$ & $\mbb{R}$ \\
          \bottomrule
        \end{tabular}
      }

      \begin{tabular}{cc}
        \\[20pt]
      \end{tabular}
    \end{center}
    \caption{Regression problem}
  \end{subfigure}
  ~
  \begin{subfigure}[b]{0.375\textwidth}
    \scalebox{0.95}{
%
\forestset{
  BDT/.style={
    for tree={
      l=1.5cm,s sep=1.15cm,
      if n children=0{}{circle}, 
      draw=midblue,
      text=midblue,
      edge={
        my edge
      },
      edge=thick,
    }
  },
}
\begin{forest}
  BDT
  [{$x_1$}, label={[yshift=-6.875ex]{{\tiny1}}} 
    [{$x_3$}, label={[yshift=-6.875ex]{{\tiny2}}}, 
      edge label={node[midway,left,xshift=-0.5pt] {{\scriptsize$\in\{0\}$}}}
      [{$x_2$}, label={[yshift=-6.875ex]{{\tiny4}}}, 
        edge label={node[midway,left,xshift=-1.5pt] {{\scriptsize$\in\{1\}$}}}
        [\dghlight{$\mathbf{\sfrac{9}{2}}$}, label={[yshift=-5.95ex]{{\tiny6}}},
          edge label={node[midway,left,xshift=-0.5pt] {{\scriptsize$\in\{0\}$}}}, rectangle, fill={tblue2!25} ]
        [\dghlight{$\mathbf{\sfrac{9}{4}}$}, label={[yshift=-5.95ex]{{\tiny7}}},
          edge label={node[midway,right,xshift=-0.575pt] {{\scriptsize$\in\{1\}$}}}, rectangle, fill={tblue2!25} ]
      ]
      [\dghlight{\textbf{0}}, label={[yshift=-5.25ex]{{\tiny5}}},
        edge label={node[midway,right,xshift=-0.5pt] {{\scriptsize$\in\{0,2\}$}}},
        rectangle, fill={tblue2!20} ]
    ]
    [\dghlight{$\mathbf{\sfrac{1}{2}}$}, label={[yshift=-5.95ex]{{\tiny3}}},
      edge={very thick,draw=purple3}, edge label={node[midway,right,xshift=0.5pt] {{\scriptsize$\in\{1\}$}}},
      rectangle, fill={tblue2!25} ]
  ]
\end{forest}
    }

    \begin{tabular}{cc}
      \\[0pt]
    \end{tabular}

    \caption{Regression tree, defining $\rho$}
  \end{subfigure}
  ~
  \begin{subfigure}[b]{0.2575\textwidth}
    \begin{center}
      \scalebox{0.95}{
        \renewcommand{\arraystretch}{1.0}
        \begin{tabular}{ccccc} \toprule
          row & $x_1$ & $x_2$ & $x_3$ & $\rho(\mbf{x})$ \\ \toprule
          1 & 0 & 0 & 0 & 0 \\
          2 & 0 & 0 & 1 & $\sfrac{9}{2}$ \\
          3 & 0 & 0 & 2 & 0 \\
          4 & 0 & 1 & 0 & 0 \\
          5 & 0 & 1 & 1 & $\sfrac{9}{4}$ \\
          6 & 0 & 1 & 2 & 0 \\
          7 & 1 & 0 & 0 & $\sfrac{1}{2}$ \\
          8 & 1 & 0 & 1 & $\sfrac{1}{2}$ \\
          9 & 1 & 0 & 2 & $\sfrac{1}{2}$ \\
          10 & 1 & 1 & 0 & $\sfrac{1}{2}$ \\
          11 & 1 & 1 & 1 & $\sfrac{1}{2}$ \\
          12 & 1 & 1 & 2 & $\sfrac{1}{2}$ \\
          \bottomrule
        \end{tabular}
      }
    \end{center}
    \caption{Tabular form}
  \end{subfigure}
  \caption{Example regression model, adapted
    from~\cite{msh-cacm24}. The sample is $((1,1,2),\sfrac{1}{2})$.
  }
  \label{fig:runex01}
\end{figure*}

\paragraph{A running example.}
%
\cref{fig:runex01} shows a regression tree, which will be used as the 
running example throughout the paper. (Given the focus on
classification models of past work on logic-based
explainability~\cite{msi-aaai22,ms-rw22,darwiche-lics23}, we opt to
study instead a regression model; this serves to illustrate that
logic-based explainability can be used both with classification and
with regression models.)
Similarly to earlier work~\cite{iims-corr20,iims-jair22}, each node in
the RT is numbered, so that paths can be easily referred to. For
example the path yielding prediction $0$ is $\langle1,2,5\rangle$.

\paragraph{Additional notation.}
%
%
An explanation problem is a tuple $\fml{E}=(\fml{M},(\mbf{v},q))$,
where $\fml{M}$ can either be a classification or a regression model,
and $(\mbf{v},q)$ is a target sample, with
$\mbf{v}\in\mbb{F}$.
(Observe that $q=\kappa(\mbf{v})$, with
$q\in\fml{K}$, in the case of a classification model, and  
$q=\rho(\mbf{v})$, with $q\in\mbb{V}$, in the case of a regression
model.)
%
%
%
Given $\mbf{x},\mbf{v}\in\mbb{F}$, and $\fml{S}\subseteq\fml{F}$, the
predicate $\mbf{x}_{\fml{S}}=\mbf{v}_{\fml{S}}$ is defined as follows:
\[
\mbf{x}_{\fml{S}}=\mbf{v}_{\fml{S}} ~~ := ~~ \left(\bigwedge\nolimits_{i\in\fml{S}}x_i=v_i\right)
\]
The set of points for which $\mbf{x}_{\fml{S}}=\mbf{v}_{\fml{S}}$
holds true is defined by
$\Upsilon(\fml{S};\mbf{v})=\{\mbf{x}\in\mbb{F}\,|\,\mbf{x}_{\fml{S}}=\mbf{v}_{\fml{S}}\}$.

\paragraph{Distributions, expected value.}
Throughout the paper, it is assumed a \emph{uniform probability
distribution} on each feature, and such that all features are
independent.
%
Thus, the probability of an arbitrary point in feature space
becomes:
%
\begin{equation}
  \prob(\mbf{x}) := \sfrac{1}{\Pi_{i\in\fml{F}}|\mbb{D}_i|}
\end{equation}
That is, every point in the feature space has the same probability.
The \emph{expected value} of an ML model $\tau:\mbb{F}\to\mbb{T}$
is denoted by $\mbf{E}[\tau(\mbf{x}]$. 
%
Furthermore, let
$\exv[\tau(\mbf{x})\,|\,\mbf{x}_{\fml{S}}=\mbf{v}_{\fml{S}}]$
represent the expected of $\tau$ over points in feature space
consistent with the coordinates of $\mbf{v}$ dictated by $\fml{S}$,
which is defined as follows:
%
%
\begin{equation} \label{eq:evdef}
  \exv[\tau(\mbf{x})\,|\,\mbf{x}_{\fml{S}}=\mbf{v}_{\fml{S}}]
  :=\sfrac{1}{|\Upsilon(\fml{S};\mbf{v})|}
  \sum\nolimits_{\mbf{x}\in\Upsilon(\fml{S};\mbf{v})}\tau(\mbf{x})
\end{equation}
%
%
Similarly, we define,
\begin{equation} \label{eq:probdef}
  \prob(\pi(\mbf{x})\,|\,\mbf{x}_{\fml{S}}=\mbf{v}_{\fml{S}}):=
  \sfrac{1}{|\Upsilon(\fml{S};\mbf{v})|}
  \sum\nolimits_{\mbf{x}\in\Upsilon(\fml{S};\mbf{v})}\tn{ITE}(\pi(\mbf{x}),1,0)
\end{equation}
where $\pi:\mbb{F}\to\{0,1\}$ is some predicate. 

\jnoteF{
  Given $\mbf{z}\in\mbb{F}$ and$\fml{S}\subseteq\fml{F}$, let
  $\mbf{z}_{\fml{S}}$ represent the vector composed of the coordinates
  of $\mbf{z}$ dictated by $\fml{S}$.
  \[
  \exv{\kappa\,|\,\mbf{x}_{\fml{S}}=\mbf{v}_{\fml{S}}}
  :=\frac{1}{|\Upsilon(\fml{S};\mbf{v})|}
  \sum\nolimits_{\mbf{x}\in\Upsilon(\fml{S};\mbf{v})}\kappa(\mbf{x})
  \]
}

\paragraph{Shapley values.}
%
Shapley values were proposed in the context of game theory in the
early 1950s by L.\ S.\ Shapley~\cite{shapley-ctg53}. Shapley values
were defined given some set $\fml{S}$, and a \emph{characteristic
function}, i.e.\ a real-valued function defined on the subsets of
$\fml{S}$, $\cf:2^{\fml{S}}\to\mbb{R}$.%
\footnote{%
The original formulation also required super-additivity of the
characteristic function, but that condition has been relaxed in more
recent works~\cite{dubey-ijgt75,young-ijgt85}.}.
It is well-known that Shapley values represent the \emph{unique}
function that, given $\fml{S}$ and $\cf$, respects a number of
important axioms. More detail about Shapley values is available in
standard
references~\cite{shapley-ctg53,dubey-ijgt75,young-ijgt85,roth-bk88}.

\paragraph{SHAP scores.} \label{par:svs}
%
In the context of explainability, Shapley values are most often
referred to as SHAP scores%
~\cite{kononenko-jmlr10,kononenko-kis14,lundberg-nips17,barcelo-aaai21,barcelo-jmlr23},
and consider a specific characteristic function
$\cf_e:2^{\fml{F}}\to\mbb{R}$,
which is defined by,
\begin{equation} \label{eq:cfs}
  \cf_e(\fml{S};\fml{E}) :=
  \exv[\kappa(\mbf{x})\,|\,\mbf{x}_{\fml{S}}=\mbf{v}_{\fml{S}}]
\end{equation}
%
%
%
Thus, given a set $\fml{S}$ of features,
$\cf_e(\fml{S};\fml{E})$ represents the \emph{e}xpected value
of the classifier over the points of feature space represented by
$\Upsilon(\fml{S};\mbf{v})$.
%
The formulation presented in earlier
work~\cite{barcelo-aaai21,barcelo-jmlr23} allows for different input
distributions when computing the average values. For the purposes of
this paper, it suffices to consider solely a uniform input 
distribution, and so the dependency on the input distribution is not
accounted for.
Independently of the distribution considered, it should be clear that
in most cases $\cfn{e}(\emptyset)\not=0$; this is the case for example
with boolean classifiers~\cite{barcelo-aaai21,barcelo-jmlr23}.


\begin{exmpl}
  We compute the value of the characteristic function $\cfn{e}$ for
  all the possible values of $\fml{S}$, i.e.\ the possible sets of
  fixed features. This tantamounts to computing the expected values of
  the regression model by fixing some of the features.
  \begin{figure*}[t]
    \begin{minipage}{0.45\textwidth}
      \begin{center}
        \renewcommand{\arraystretch}{1.125}
        \renewcommand{\tabcolsep}{0.75em}
        \begin{tabular}{cc} \toprule
          $\fml{S}$ & $\cfn{e}(\fml{S};\fml{E})$ \\ \toprule
          $\emptyset$ & $\sfrac{13}{16}$ \\
          $\{1\}$ & $\sfrac{1}{2}$ \\
          $\{2\}$ & $\sfrac{5}{8}$ \\ 
          $\{3\}$ & $\sfrac{1}{4}$ \\
          \bottomrule
        \end{tabular}
      \end{center}
    \end{minipage}
    \begin{minipage}{0.45\textwidth}
      \begin{center}
        \renewcommand{\arraystretch}{1.125}
        \renewcommand{\tabcolsep}{0.75em}
        \begin{tabular}{cc} \toprule
          $\fml{S}$ & $\cfn{e}(\fml{S};\fml{E})$ \\ \toprule
          $\{1,2\}$ & $\sfrac{1}{2}$ \\
          $\{1,3\}$ & $\sfrac{1}{2}$ \\
          $\{2,3\}$ & $\sfrac{1}{4}$ \\
          $\{1,2,3\}$ & $\sfrac{1}{2}$ \\
          \bottomrule
        \end{tabular}
      \end{center}
    \end{minipage}
    %
    \caption{Expected values, for all possible sets of fixed features}
    \label{fig:cfs01}
  \end{figure*}
  The resulting values are shown in~\cref{fig:cfs01}.
\end{exmpl}

To simplify the notation, the following definitions are used,
\begin{align}
  \Delta_i(\fml{S};\fml{E},\cf) & :=
  \left(\cf(\fml{S}\cup\{i\})-\cf(\fml{S})\right)
  \label{eq:def:delta}
  \\[2pt] 
  \varsigma(|\fml{S}|) & :=
  \sfrac{|\fml{S}|!(|\fml{F}|-|\fml{S}|-1)!}{|\fml{F}|!} 
  \label{eq:def:vsigma}
\end{align}
(Observe that $\Delta_i$ is parameterized on $\fml{E}$ and $\cf$.)

Finally, let $\svn{E}:\fml{F}\to\mbb{R}$, i.e.\ the SHAP score for
feature $i$, be defined by,\footnote{%
Throughout the paper, the definitions of $\Delta_i$ and $\sv$ are
explicitly associated with the characteristic function used in their
definition.}
\begin{equation} \label{eq:sv}
  \svn{E}(i;\fml{E},\cfn{e}):=\sum\nolimits_{\fml{S}\subseteq(\fml{F}\setminus\{i\})}\varsigma(|\fml{S}|)\times\Delta_i(\fml{S};\fml{E},\cfn{e}) 
\end{equation}
Given a sample $(\mbf{v},q)$, the SHAP score assigned to each
feature measures the \emph{contribution} of that feature with respect
to the prediction. 
From earlier work, it is understood that a positive/negative value
indicates that the feature can contribute to changing the prediction,
whereas a value of 0 indicates no
contribution~\cite{kononenko-jmlr10}.
%

%

%

%

\jnoteF{%
  Feature selection vs.\ feature attribution.\\
  Intrinsic interpretability.\\
  Model-agnostic XPs.
}

\jnoteF{More XAI surveys!!!!}

\paragraph{A brief peek at explanability.}
We now briefly overview approaches to XAI that are not logic-based.
As clarified later in the paper, all of this work exhibits important
misconceptions. As a result, computed explanations can lack in 
rigor and can induce human decision-makers in error.%
\footnote{
For the interested reader, there is a wealth of resources on
non-rigorous methods of
XAI~\cite{gunning-darpa16,muller-dsp18,pedreschi-acmcs19,gunning-aimag19,gunning-sr19,xai-bk19,muller-xai19-ch01,molnar-bk20,muller-ieee-proc21,molnar-bk23,mishra-bk23}.}

Given an explanation problem, XAI by \emph{feature attribution} seeks
to assign relative importance to the features of the ML model.
Two general-purpose methods have been
devised~\cite{guestrin-kdd16,lundberg-nips17}, but there exist also
methods specific two explaining neural networks
(NNs)~\cite{muller-plosone15}.
Among other examples, explanations for NNs can be based on adapting
backpropagation for propagating information about a prediction across
the layers of the NN~\cite{muller-plosone15}.
LIME~\cite{guestrin-kdd16} approximates a complex ML model with a
linear model, by finding the coefficients to assign to each feature in
the linear model. SHAP~\cite{lundberg-nips17} finds its theoretical
underpinnings in game theory, more concretely on Shapley values, and
can be related with earlier work on explaining ML
models~\cite{kononenko-jmlr10,kononenko-kis14}.
Both LIME and SHAP are model-agnostic, i.e.\ the actual implementation
of the ML model is ignored when computing explanations.
At present, SHAP ranks among the most successful XAI
approaches~\cite{molnar-bk23,mishra-bk23}.


An alternative to assigning a score of importance to each feature, is
\emph{feature selection}. Given an explanation problem, feature selection
identifies a set of features which explain the prediction. One of the
best-known approache for feature selection is
Anchors~\cite{guestrin-aaai18}.
Similarly to LIME and SHAP, Anchors is model-agnostic.

\section{The Recent Past: Shaping Logic-Based Explainability}
\label{sec:past}

The study of explainability in AI can be traced back several
decades, starting the in 1970s and
1980s~\cite{swartout-ijcai77,swartout-aij83,shanahan-ijcai89}, but
continuing over the
years~\cite{simari-aij02,marquis-aij02,uzcategui-aij03}.
Furthermore, the links of logic-based abduction with explanations are
also well-established~\cite{gottlob-jacm95}.
The purpose of this section is to provide an account of the recent
work on logic-based explainability for the specific purpose of
explaining ML models.

It should be noted that logic-based explanations of ML models have
been approache both from the perspective of logic-based
abduction~\cite{inms-aaai19} and as prime implicants of boolean
functions~\cite{darwiche-ijcai18}.
The two perspectives are equivalent, and we opt for adopting the
connection with logic-based abduction, since it offers a more general
framework, which can encompass non-boolean classifiers, regression
models, among others.

%
%

\subsection{Defining Logic-Based Explanations}

\jnoteF{AXps, CXps, duality.}

\jnoteF{It is trivial to realize that, in the case of plain
  classifiers,  CXps represent \emph{actionable recourse}.}

This section briefly overviews formal explanations, one example of
logic-based explainability that has been studied in recent
years~\cite{msi-aaai22,ms-rw22,darwiche-lics23}.
Nevertheless, we propose a simple extension that also enables
accounting for regression problems, in addition to classification
problems.

\paragraph{Similarity predicate.}
%
Similarity functions have been used in explainable AI in recent
work~\cite{lhms-corr24,lhams-corr24} to account for feature
importance. We adapt this concept for the purpose of explainability by
feature selection.

Given an ML model and some input $\mbf{x}$, the output of the ML model
is \emph{distinguishable} with respect to the sample $(\mbf{v},q)$ if the
observed change in the model's output is deemed sufficient; 
otherwise it is \emph{similar} (or indistinguishable).
This is represented by a \emph{similarity} predicate (which can be
viewed as a boolean function) 
$\similar:\mbb{F}\to\{\bot,\top\}$ (where $\bot$ signifies
\emph{false}, and $\top$ signifies \emph{true}). Concretely,
$\similar(\mbf{x};\fml{E})$ holds true iff the change in the ML model
output is deemed \emph{insufficient} and so no observable difference
exists between the ML model's output for $\mbf{x}$ and $\mbf{v}$.%
\footnote{
Throughout the paper, parameterization are shown after the separator
';', and will be elided when clear from the context.}
For regression problems, we write instead $\similar$ as the
instantiation of a template predicate,
i.e.\ $\similar(\mbf{x};\fml{E})=\tsimilar(\mbf{x};\fml{E},\delta)$,
where $\delta$ is an optional measure of output change, which can be
set to 0.%
\footnote{%
Exploiting a threshold to decide whether there exists an observable
change has been used in the context of adversarial
robustness~\cite{barrett-nips23}. Furthermore, the relationship
between adversarial examples and explanations is
well-known~\cite{inms-nips19,barrett-nips23}.}

For regression problems, we represent relevant changes to the output
by a parameter $\delta$. Given a change in the input from $\mbf{v}$ to
$\mbf{x}$, a change in the output is indistinguishable (i.e.\ the
outputs are similar) if,
\[
\similar(\mbf{x};\fml{E}) 
:= \tsimilar(\mbf{x};\fml{E},\delta)
:= [|\rho(\mbf{x})-\rho(\mbf{v})|\le\delta]
\]
otherwise, it is distinguishable.

For classification problems, similarity is defined to equate with not
changing the predicted class. Given a change in the input from 
$\mbf{v}$ to $\mbf{x}$, a change in the output is indistinguishable
(i.e.\ the outputs are similar) if,
\[
\similar(\mbf{x};\fml{E}):=[\kappa(\mbf{x})=\kappa(\mbf{v})]
\]
otherwise, it is distinguishable. (As shown in the remainder of this
paper, $\similar$ allows abstracting away whether the underlying
model implements classification or regression.)

\begin{exmpl}
  For the running example, we will pick
  $\delta=\sfrac{1}{2}$ and define 
  the similarity predicate as follows
  $\similar(\mbf{x};\fml{E}):=\tsimilar(\mbf{x};\fml{E},\sfrac{1}{2})$.
  %
  We note that the choice of $\delta$ is such that the computation of
  logic-based explanations is still straighforward.
\end{exmpl}

The definition of the similarity predicate can also integrate more
general definitions of explanations, including inflated
explanations~\cite{iisms-aaai24} and explanations with
partially-specified inputs and predicting sets of
classes~\cite{bmpms-corr23}.

\paragraph{Abductive and contrastive explanations (AXps/CXps).}
AXps and CXps are examples of formal
explanations for classification
problems~\cite{kutyniok-jair21,msi-aaai22,darwiche-lics23}. We propose
a generalization that encompases regression problems.

A weak abductive explanation (WAXp) denotes a set of features
$\fml{S}\subseteq\fml{F}$, such that for every point in feature space
the ML model output in similar to the given sample: $(\mbf{v},q)$.
The condition for a set of features $\fml{S}\subseteq\fml{F}$ to
represent a WAXp (which also defines a corresponding predicate
$\waxp$) is as follows:
\[
\waxp(\fml{S}) ~~ := ~~
\forall(\mbf{x}\in\mbb{F}).(\mbf{x}_{\fml{S}}=\mbf{v}_{\fml{S}})\limply\similar(\mbf{x};\fml{E})
\]
Alternatively, we write,
\[
\waxp(\fml{S};\fml{E}) ~~ := ~~
\prob(\similar(\mbf{x};\fml{E})\,|\,\mbf{x}_{\fml{S}}=\mbf{v}_{\fml{S}})
= 1
\]
(The definition of WAXps using probabilities is of interest when exact
computation is impractical.)
Furthermore, an AXp is a subset-minimal WAXp.

A weak contrastive explanation (WCXp) denotes a set of features
$\fml{S}\subseteq\fml{F}$, such that there exists some point in
feature space, where only the features in $\fml{S}$ are allowed to
change, that makes the ML model output distinguishable from the given
sample $(\mbf{v},q)$.
The condition for a set of features $\fml{S}\subseteq\fml{F}$ to
represent a WCXp (which also defines a corresponding predicate
$\wcxp$) is as follows:
\[
\wcxp(\fml{S};\fml{E}) ~~ := ~~
\exists(\mbf{x}\in\mbb{F}).(\mbf{x}_{\fml{F}\setminus\fml{S}}=\mbf{v}_{\fml{F}\setminus\fml{S}})\land\neg\similar(\mbf{x};\fml{E},\delta)
\]
Alternatively, we write,
\[
\wcxp(\fml{S};\fml{E}) ~~ := ~~
\prob(\similar(\mbf{x};\fml{E})\,|\,\mbf{x}_{\fml{F}\setminus\fml{S}}=\mbf{v}_{\fml{F}\setminus\fml{S}})
< 1
\]
(As before, the definition of WCXps using probabilities is of interest
when exact computation is impractical.)
Furthermore, a CXp is a subset-minimal WCXp.
Observe that for ML models that compute some function, (W)CXps
can be viewed as implementing actionable
recourse~\cite{spangher-fat19}. Furthermore, (W)CXps formalize the
notion of contrastive explanation discussed
elsewhere~\cite{miller-aij19}.


\begin{figure*}[t]
  \begin{subfigure}[b]{\textwidth} 
    \centering
    \renewcommand{\tabcolsep}{0.8725em}
    \begin{tabular}{cccc}
      \toprule[1pt]
      $\fml{S}$ &
      $\rows(\fml{S})$ &
      \makecell{$\waxp(\fml{S})$?\\$\fml{S}$ sufficient?} &
      \makecell{$\axp(\fml{S})$?\\$\fml{S}$ also minimal?} 
      %
      \\
      \midrule[0.875pt]
      $\emptyset$ &
      1..12 & \nomark & 
      \\
      $\{1\}$ &
      7,8,9,10,11,12 & \yesmark & \yesmark 
      \\
      $\{2\}$ &
      4,5,6,10,11,12 & \nomark & 
      \\
      $\{3\}$ &
      3,6,9,12 & \nomark & 
      \\
      $\{1,2\}$ & 10,11,12 & \yesmark & \nomark 
      \\
      $\{1,3\}$ & 9,12 & \yesmark & \nomark 
      \\
      $\{2,3\}$ & 6,12 & \nomark & 
      \\
      $\{1,2,3\}$ & 12 & \yesmark & \nomark 
      \\
      \bottomrule[1pt]
    \end{tabular}
    \caption{Computation of AXps; $\fml{S}$ represents the candidate AXp}
  \end{subfigure}

  \medskip

  \begin{subfigure}[b]{\textwidth}
    \centering
    \renewcommand{\tabcolsep}{0.795em}
    \begin{tabular}{cccc}
      \toprule[1pt]
      %
      $\fml{F}\setminus\fml{S}$ &
      $\rows(\fml{F}\setminus\fml{S})$ &
      \makecell{$\wcxp(\fml{S})$?\\$\fml{S}$ changes $\kappa$?} &
      \makecell{$\cxp(\fml{S})$?\\$\fml{S}$ also minimal?}
      \\
      \midrule[0.875pt]
      %
      $\{1,2,3\}$ & 12 & \nomark & 
      \\
      %
      $\{2,3\}$ & 6,12 & \yesmark & \yesmark
      \\
      %
      $\{1,3\}$ & 9,12 & \nomark &
      \\
      %
      $\{1,2\}$ & 10,11,12 & \nomark &
      \\
      %
      $\{3\}$ & 3,6,9,12 & \yesmark & \nomark
      \\
      %
      $\{2\}$ & 4,5,6,10,11,12 & \yesmark & \nomark
      \\
      %
      $\{1\}$ & 7,8,9,10,11,12 & \nomark & 
      \\
      %
      $\emptyset$ & 1..12 & \yesmark & \nomark
      \\
      \bottomrule[1pt]
    \end{tabular}
    \caption{Computation of CXps; $\fml{S}$ (not shown) represents the
      candidate CXp}
  \end{subfigure}
  \caption{Computing AXps/CXps for the running example
    of~\cref{fig:runex01} and sample
    $(\mbf{v},c)=((1,1,2),\sfrac{1}{2})$. All subsets of features are 
    considered.
    For computing AXps/CXps, and for some set $\fml{Z}$, the features in
    $\fml{Z}$ are fixed to their values as determined by $\mbf{v}$.
    The picked rows, i.e.\ $\rows(\fml{Z})$, are the rows consistent
    with those fixed values.
  }
  \label{fig:xps01}
\end{figure*}

\begin{exmpl}
  For the running example, it is plain that:
  \begin{enumerate}[nosep]
  \item If the value of $x_1$ is 1, then the value of
    $\rho(1,x_2,x_3)$ is guaranteed to be $\sfrac{1}{2}$;
  \item Otherwise, if the value of $x_1$ is 0 then,
    (i) if the value of $x_3\in{0,2}$, then the value of
    $\rho(0,x_2,x_3)=0$;
    (ii) if the value of $x_3=1$, then we have that
    $\rho(0,0,1)=\sfrac{9}{2}$ and $\rho(0,1,1)=\sfrac{9}{4}$.
  \end{enumerate}
  Thus, given the sample $((1,1,2),\sfrac{1}{2})$,  if we fix $x_1$,
  then the prediction does not change, no matter the values assigned
  to $x_2$ and $x_3$. More formally, we can state that it holds true
  that
  $\forall(x_2\in\{0,1\})\forall(x_3\in\{0,1,2\}).\similar((1,x_2,x_3);\fml{E})$.
  Also, to change the prediction to a value other
  that $\sfrac{1}{2}$, it suffices to change the value of
  $x_1$.
  Once again, we can state more formally that it holds true that
  $\neg\similar((0,1,2);\fml{E})$.
  As a result, it is the case that there is only on AXp,
  i.e.\ $\{1\}$, and a single CXp, i.e.\ $\{1\}$.\\
  The computation of AXps/CXps is summarized in~\cref{fig:xps01}.
\end{exmpl}

Finally, it is well-known that a set of features is an AXp iff it is a
minimal hitting set of the set of CXps, and
vice-versa~\cite{inams-aiia20}. (This result has been proved for
classification problems, but the framework detailed earlier in the
paper generalizes the result also to regression problems.)

\paragraph{Relationship with adversarial examples.}
%
Adversarial examples serve to reveal the brittleness of ML
models~\cite{szegedy-iclr14,szegedy-iclr15}. Adversarial robustness
aims to attest to the absence of adversarial examples on
representative inputs. The importance of deciding adversarial
robustness is illustrated by a wealth of competiting
alternatives~\cite{johnson-sttt23}.

Given a sample $(\mbf{v},q)$, and a norm $l_p$,%
\footnote{The definition of norm $l_p$, $p\ge0$, is standard in ML: 
\url{https://en.wikipedia.org/wiki/Norm_(mathematics)}. For the
purposes of robustness in ML, examples of norms considered include
$l_0$ (Hamming distance), $l_1$ (Manhattan distance), $l_2$
(Euclidean distance) and $l_{\infty}$ (Chebyshev distance).}
a point $\mbf{x}\in\mbb{F}$ is an \emph{adversarial example} if the
prediction for $\mbf{x}$ is distinguishable from that for
$\mbf{v}$. Formally, we write,
\[
\aex(\mbf{x};\fml{E}) ~~ := ~~
\left(||\mbf{x}-\mbf{v}||_{p}\le\epsilon\right)\land
\neg\similar(\mbf{x};\fml{E})
\]
where the $l_p$ distance between the given point $\mbf{v}$ and other
points of interest is restricted to $\epsilon>0$.
%
%
Moreover, we define a \emph{constrained} adversarial example, such
that the allowed set of points is given by the predicate
$\mbf{x}_{\fml{S}}=\mbf{v}_{\fml{S}}$. Thus,
\[
\aex(\mbf{x},\fml{S};\fml{E}) ~~ := ~~
\left(||\mbf{x}-\mbf{v}||_{p}\le\epsilon\right)\land
\left(\mbf{x}_{\fml{S}}=\mbf{v}_{\fml{S}}\right)\land
\neg\similar(\mbf{x};\fml{E})
\]
Adversarial robustness is concerned with assessing whether complex ML
models do not exhibit adversarial examples for chosen samples.
Given the previous definitions, we get the following result.
\begin{prop}
  $\exists(\mbf{x}\in\mbb{F}).\left(\aex(\mbf{x},\fml{F}\setminus\fml{S};\fml{E})\right)$
  iff
  $\wcxp(\fml{S};\fml{E})$, i.e.\ there exists a constrained
  adversarial example with the features $\fml{F}\setminus\fml{S}$ iff
  the set $\fml{S}$ is a WCXp. 
\end{prop}

\paragraph{Feature (ir)relevancy.}
%
The set of features that are included in at least one (abductive) 
explanation are defined as follows:
\begin{equation}
  \mathfrak{F}(\fml{E}):=\{i\in\fml{X}\,|\,\fml{X}\in2^{\fml{F}}\land\axp(\fml{X})\}
\end{equation}
%
(A well-known result is that $\mathfrak{F}(\fml{E})$ remains unchanged
if CXps are used instead of AXps~\cite{ms-rw22,cms-aij23}, in which
case predicate $\cxp(\fml{X})$ holds true iff $\fml{X}$ is a CXp.) 
Finally, a feature $i\in\fml{F}$ is \emph{irrelevant}, i.e.\ predicate
$\irrelevant(i)$ holds true, if $i\not\in\mathfrak{F}(\fml{E})$;
otherwise feature $i$ is \emph{relevant}, and predicate $\relevant(i)$
holds true. 
Clearly, given some explanation problem $\fml{E}$,
$\forall(i\in\fml{F}).\irrelevant(i)\leftrightarrow\neg\relevant(i)$.

\begin{exmpl}
  For the running example, it is clear that feature 1 is relevant, and
  that features 2 and 3 are irrelevant.
\end{exmpl}



\subsection{Complexity of Computing Explanations}
\label{ssec:cplxp}


%

\paragraph{Tractable cases.}
%
The progress observed in logic-based explainability in recent years
enabled proving that the computation of one single AXp or CXp is
tractable for several families of classifiers. These include
decision trees and graphs~\cite{iims-corr20,hiims-kr21,iims-jair22},
naive bayes classifiers~\cite{msgcin-nips20},
monotone classifiers~\cite{msgcin-icml21},
d-DNNF classifiers~\cite{hiicams-aaai22},
among several other examples~\cite{cms-cp21,cms-aij23,ccms-kr23}.


\jnoteF{Add example(s)...}

\begin{table}[t]
  \renewcommand{\tabcolsep}{0.5em}
  \renewcommand{\arraystretch}{1.15}
  \begin{tabular}{lccl} \toprule
    Current $\fml{S}$ &
    Dropped feature &
    Path consistent with $\neg\similar(\mbf{x};\fml{E})$? & 
    Resulting $\fml{S}$
    \\ \toprule
    $\{1,2,3\}$ &
    1 &
    \yesmark, $\langle1,2,5\rangle$ &
    $\{1,2,3\}$
    \\
    $\{1,2,3\}$ &
    2 &
    \nomark &
    $\{1,3\}$
    \\
    $\{1,3\}$ &
    3 &
    \nomark &
    $\{1\}$ --- AXp
    \\
    \bottomrule
  \end{tabular}
  \caption{Computation of one AXp. $\fml{S}$ denotes the set of fixed
    features. A dropped feature is allowed to take any value from its
    domain.} \label{tab:axp01}
\end{table}

\begin{example}
  \cref{tab:axp01} summarizes the execution of a polynomial-time
  algorithm for computing one AXp for a
  RT~\cite{iims-corr20,iims-jair22}.  For each feature $i$ (in any
  order), we allow the feature $i$ to take any value from its domain.
  Afterwards,  we check the paths of the RT, with leaves predicting a
  value that is distinguished from the prediction $q$. If any of those
  paths can be made consistent, by picking suitable values for the
  free features, then the feature $i$ must be fixed (and so added back
  to $\fml{S}$. Otherwise, we can safely drop the feature $i$. After
  analyzing all features, the resulting set $\fml{S}$ is an
  AXp. (Different algorithms can be envisioned; here we adopted one of
  the algorithms detailed in~\cite{iims-jair22}.)
\end{example}


%

\paragraph{Intractable cases.}
%
Besides families of classifiers for which computing one AXp or CXp is
tractable, recent work also established the intractability (in terms
of computational complexity) of computing one AXp/CXp for other
families of classifiers.%
\footnote{%
As argued in recent work~\cite{ms-rw22}, intractability from the
viewpoint of computational complexity does not necessarily equate with
practical intractability, given the progress observed in automated
reasoning over the last couple of decades.}
These include
decision lists~\cite{ims-sat21},
random forests~\cite{ims-ijcai21} and other tree
ensembles~\cite{iisms-aaai22}.%
\footnote{
More recent
work~\cite{marquis-aaai22,marquis-ijcai22a,marquis-ijcai22b,marquis-aistats23} 
proposed some improvements to the core ideas for explaining random
forest and tree ensembles.}



\paragraph{Tackling intractability with automated reasoning.}
%
In the cases of families of classifiers for which computing one
explanation is computationally hard, one of the main challenges of
logic-based XAI has been the development of logic encodings that
enable the efficient integration of automated reasoners.
For example, for decision lists and random forests (with majority
voting) it has been shown that propositional encodings suffice, which
in turn enables the use of highly optimized SAT
solvers~\cite{ims-sat21,ims-ijcai21}, based on the conflict-driven
clause learning
paradigm~\cite{mss-iccad96,mss-tcomp99,msm-hbk18,mslm-hbk21}.
In the case of boosted trees and other tree ensembles, it was possible
to devised first SMT-based encodings~\cite{inms-corr19} and more
recently the use of Maximum Satisfiability (MaxSAT)
solvers~\cite{iisms-aaai22}.
Despite the ability to efficiently explain decision lists, random
forest and other tree ensembles, finding explanations for medium to
large-size neural networks has been a challenge since
2019~\cite{inms-aaai19}.

%
%

\paragraph{Tackling intractability with knowledge compilation.}
%
When computing AXps/CXps, the main alternative to the use of automated
reasoners is to compile a tractable representation, from which
explanation can be efficiently
computed~\cite{darwiche-ijcai18,darwiche-aaai19,darwiche-ecai20,darwiche-pods20,darwiche-lics23,darwiche-jlli23}.
Despite the large body of work, the size and complexity of the models
that can be efficiently explained remains a major drawback to the use
of compilation-based approaches.

\jnoteF{Compilation-based approaches.
  The belief is that complex models can be compiled into some sort
  tractable representation. Despite the important theoretical advances
  in KR\&R and KC, examples of practical successes are scarce. Also,
  data points in ML are worrisome (see Bertossi).
}

%
%
%
%

\section{The Present: Expanding Logic-Based Explainability} 
\label{sec:present}

\subsection{Explainability Queries}



Besides the computation of rigorous explanations, research in
logic-based explainability has studied a growing number of queries of
interest~\cite{inams-aiia20,marquis-kr20,hiims-kr21,marquis-kr21}. We
briefly overview the enumeration of explanations~\cite{inams-aiia20}
and deciding feature relevancy and necessity.

For most families of ML models, enumeration of explanations hinges of
the MARCO algorithm~\cite{lpmms-cj16} for the enumeration of MUSes and
MCSes. As shown in several earlier works, minor modifications to the
MARCO algorithm enable the enumeration of both AXps and
CXps~\cite{inams-aiia20,ms-rw22}.
Nevertheless, polynomial-delay enumeration of AXps was proved in the
case of NBCs~\cite{msgcin-nips20}.

The notions of relevancy and necessity in abduction can be traced to
the earlier works on logic-based abduction~\cite{gottlob-jacm95}.
A feature is relevant if it occurs in some explanation. In contrast, a
feature is necessary if it occurs in all explanations.
A number of recent works have studied both feature relevancy and
necessity\cite{hiims-kr21,hims-aaai23,hcmpms-tacas23}, proposing
practically efficient algorithms based on abstraction refinement.

\subsection{Explanation Size}


The cognitive limits of human decision-makers are
well-known~\cite{miller-pr56}. Unfortunately, AXps/CXps can in some
domains be significantly larger than what a human decision-maker can
understand.
Logic-based explainability has proposed to replace the exact
definition of AXp by approximate definitions, while still providing
strong theoretical
guarantees~\cite{kutyniok-jair21,iisncms-corr21,ims-corr22a,iincms-corr22,ihincms-corr22,barcelo-nips22,ihincms-ijar23,imms-corr23}.
Probabilistic explanations are sets of feature which guarantee that
the probabilitic of getting a prediction other than the target will be
below some threshold. Recent work~\cite{ihincms-ijar23,imms-corr23}
developed algorithms for the practically efficient computation of
probabilistic AXps for several families of classifiers.

\subsection{Expressivity of Explanations}


The definition of AXps takes into considered the values assigned to
the features, i.e.\ if a feature $i$ is part of an AXp, then we
require that the values of feature $i$ to be those specified by $v_i$,
i.e.\ we implicitly assume a literal $x_i=v_i$. However, it can be
convenient to allow for more expressive literals. The following
example illustrates, the concept of \emph{inflated explanations}
proposed in recent work~\cite{iisms-aaai24}.

\begin{exmpl}
  We consider the running example (see~\cref{fig:runex01}), but a
  different sample: $(0,1,0)$. In this case, by inspection of the RT,
  it is apparent that one AXp is $\{1,3\}$, i.e.\ if both features 1
  and 3 are fixed to their values (i.e.\ $x_1=0\land{x_3=0}$), then
  the prediction is guaranteed to be 0. This can be represented as a
  logic rule:
  \[
  \tn{IF}\quad{x_1=0}\land{x_3=0}\quad\tn{THEN}\quad\rho(\mbf{x})=\sfrac{1}{2}
  \]
  However, we can make the explanation more expressive, by replacing
  the $x_i=v_i$ literals with $x_i\in{V_i}$ literals.
  For this sample, the prediction will remain unchanged if we write
  the explanation literals as follows:
  $x_1\in\{0\}\land{x_3}\in\{0,2\}$, which can be represented with the
  following logic rule:
  \[
  \tn{IF}\quad{x_1\in\{0\}}\land{x_3\in\{0,2\}}\quad\tn{THEN}\quad\rho(\mbf{x})=\sfrac{1}{2}
  \]
  It is plain that the expanded explanation offers more information
  (and covers more points in feature space) than the original
  explanation.
\end{exmpl}

The computation of inflated AXps does not change the
complexity of computing a single plan AXp.
Besides basic expanded explanations~\cite{iisms-corr23,iisms-aaai24},
additional works have looked at variants of expanded
explanations~\cite{darwiche-corr23,darwiche-jelia23}.


\subsection{Additional Topics \& Applications}

%

Additional work in logic-based explainability covered the use of
taking into account input
constraints~\cite{rubin-aaai22,yisnms-aaai23},
exploiting surrogate models~\cite{mazure-cikm21},
and pre-computing explanations in the case of DTs (and so
RTs)~\cite{hms-ecai23}, but also initial applications in natural
language processing~\cite{kwiatkowska-ijcai21a}.

As an example, one possible criticism of logic-based explanations is
that all points in feature space can be considered. However, in many
situations, is it clear that not all inputs are realistic. One
recently proposed solution is to infer constraints on the ML model's
inputs~\cite{yisnms-aaai23}, and then compute explanations by taking
those constraints into account.

%
%

\section{By-Products: Misconceptions in XAI \& ML} \label{sec:myths}

Logic-based XAI not only offers a rigorous approach to explainability,
but it also provides a framework to uncover misconceptions with other
less-rigorous approaches to XAI. This section provides a brief account
of several misconceptions that logic-based XAI has exposed.

This section briefly overviews some of the most important
misconceptions in non-rigorous XAI, but also in adversarial
robustness. A more detailed analysis is presented
in~\cite{ms-iceccs23}.

\subsection{The Misconception of Interpretability}



Since the inception of the field of ML, it has been widely claimed
that decision trees are \emph{interpretable}. Unfortunately, the
notion of interpretability if fairly subjective, without a commonly
accepted rigorous definition~\cite{lipton-cacm18}. Nevertheless, in
recent years there are been repeated proposals regarding the use of
so-called \emph{interpretable models}, where the explanation is the
model itself~\cite{rudin-naturemi19,rudin-ss22}.
For example, in the case of decision trees (DTs), the explanation is
the tree path consistent with the target sample.
Unfortunately, it has been shown in recent years that decision trees
(DTs) can exhibit arbitrary redundancy, namely when tree paths are
used as explanations~\cite{iims-corr20}. Additional
work~\cite{hiims-kr21,iims-jair22} has provided extensive additional
evidence regarding tree paths containing arbitrarily large redundancy
(on the number of features), but also outline different algorithms for
the efficient computation of explanations. Unsurprisingly, similar
results have been obtained regarding other models often claimed to be 
interpretable~\cite{msi-frai23}.

\subsection{The Misconception of Model-Agnostic Explainability}

\jnoteF{Model-agnostic XPs are unsound, often.}

Model-agnostic explainability encompasses methods produce explanations
of ML models that are treated as black-boxes, i.e.\ explanations are
computed with reasoning about the ML model's operation.
Two well-known methods of model-agnostic explainability are
LIME~\cite{guestrin-kdd16} and SHAP~\cite{lundberg-nips17}. Both
methods are currently ubiquitously used in XAI.
Besides LIME and SHAP, both of which are examples of XAI by feature
attribution, another well-known method is
Anchors~\cite{guestrin-aaai18}, which is an example of XAI by feature
selection.

Since 2019, a growing number of results have confirmed that
model-agnostic methods of XAI will produce unsound
results~\cite{inms-corr19,ignatiev-ijcai20,yisnms-aaai23}.
Furthermore, there is evidence that model-agnostic methods not only
produce unsound results, but the produced results are far from a sound
result~\cite{nsmims-sat19}.

\subsection{Misconceptions in XAI by Feature Attribution}
\label{ssec:svflaws}

\jnoteF{SHAP scores are utterly flawed.}

\cref{sec:prelim} (see~\cpageref{par:svs}) briefly overviews the
computation of SHAP scores, which provide the theoretical
underpinnings of the tool SHAP~\cite{lundberg-nips17}.
At present, SHAP epitomizes XAI by feature attribution. Moreover, and
since its inception, SHAP has had massive impact in a wide range of
uses of XAI.
Nevertheless, over the years researchers identified issues with the
results obtained with the tool
SHAP~\cite{friedler-icml20,najmi-icml20,taly-cdmake20,guigue-icml21}.
More recently, researchers have established that even the theory
underlying SHAP scores can produce extremely unsatisfactory results,
which tantamount to misleading human
decision-makers~\cite{hms-corr23a,hms-corr23c,hms-corr23d,msh-corr23,msh-cacm24,hms-ijar24}.
We illustrate the problems with the (exact) computation of SHAP scores
using the running example.

\begin{figure*}[t]
  \begin{subfigure}{1.0\textwidth}
    \begin{center}
      \renewcommand{\tabcolsep}{0.725em}
      \begin{tabular}{cccccc}
        \toprule[1pt]
        $\fml{S}$ &
        $\phi(\fml{S})$ &
        $\phi(\fml{S}\cup\{1\})$ &
        $\Delta_1(\fml{S})$ &
        $\varsigma(\fml{S})$ &
        $\varsigma(\fml{S})\times\Delta_1(\fml{S})$
        \\
        \midrule[0.875pt]
        $\emptyset$ &
        $\sfrac{13}{16}$ &
        $\sfrac{1}{2}$ &
        $-\sfrac{5}{16}$ &
        $\sfrac{0!(3-0-1)!}{3!}=\sfrac{1}{3}$ &
        $-\sfrac{5}{48}$
        \\
        $\{2\}$ &
        $\sfrac{5}{8}$ &
        $\sfrac{1}{2}$ &
        $-\sfrac{1}{8}$ &
        $\sfrac{1!(3-1-1)!}{3!}=\sfrac{1}{6}$ &
        $-\sfrac{1}{48}$
        \\
        $\{3\}$ &
        $\sfrac{1}{4}$ &
        $\sfrac{1}{2}$ &
        $\sfrac{1}{4}$ &
        $\sfrac{1!(3-1-1)!}{3!}=\sfrac{1}{6}$ &
        $\sfrac{1}{24}$
        \\
        $\{2,3\}$ &
        $\sfrac{1}{4}$ &
        $\sfrac{1}{2}$ &
        $\sfrac{1}{4}$ &
        $\sfrac{2!(3-2-1)!}{3!}=\sfrac{1}{3}$ &
        $\sfrac{1}{12}$
        \\
        \midrule[0.75pt]
        \multicolumn{5}{r}{SHAP score for feature 1 \hfill
          $\sv(1)~~=$} &
        $0$ \\
          \bottomrule[1pt]
        \end{tabular}
      \end{center}
    \end{subfigure}

    \smallskip\smallskip\smallskip
    %
    \begin{subfigure}{1.0\textwidth}
      \begin{center}
        \renewcommand{\tabcolsep}{0.725em}
        \begin{tabular}{cccccc}
          \toprule[1pt]
          $\fml{S}$ &
          $\phi(\fml{S})$ &
          $\phi(\fml{S}\cup\{2\})$ &
          $\Delta_2(\fml{S})$ &
          $\varsigma(\fml{S})$ &
          $\varsigma(\fml{S})\times\Delta_2(\fml{S})$
          \\
          \midrule[0.875pt]
          $\emptyset$ &
          $\sfrac{13}{16}$ &
          $\sfrac{5}{8}$ &
          $-\sfrac{3}{16}$ &
          $\sfrac{0!(3-0-1)!}{3!}=\sfrac{1}{3}$ &
          $-\sfrac{3}{48}$
          \\
          $\{1\}$ &
          $\sfrac{1}{2}$ &
          $\sfrac{1}{2}$ &
          $0$ &
          $\sfrac{1!(3-1-1)!}{3!}=\sfrac{1}{6}$ &
          $0$
          \\
          $\{3\}$ &
          $\sfrac{1}{4}$ &
          $\sfrac{1}{4}$ &
          $0$ &
          $\sfrac{1!(3-1-1)!}{3!}=\sfrac{1}{6}$ &
          $0$
          \\
          $\{1,3\}$ &
          $\sfrac{1}{2}$ &
          $\sfrac{1}{2}$ &
          $0$ &
          $\sfrac{2!(3-2-1)!}{3!}=\sfrac{1}{3}$ &
          $0$
          \\
          \midrule[0.75pt]
          \multicolumn{5}{r}{SHAP score for feature 2 \hfill
            $\sv(2)~~=$} &
          $-0.0625$ \\ 
          \bottomrule[1pt]
        \end{tabular}
      \end{center}
    \end{subfigure}

    \smallskip\smallskip\smallskip
    %
    \begin{subfigure}{1.0\textwidth}
      \begin{center}
        \renewcommand{\tabcolsep}{0.725em}
        \begin{tabular}{cccccc}
          \toprule[1pt]
          $\fml{S}$ &
          $\phi(\fml{S})$ &
          $\phi(\fml{S}\cup\{3\})$ &
          $\Delta_3(\fml{S})$ &
          $\varsigma(\fml{S})$ &
          $\varsigma(\fml{S})\times\Delta_3(\fml{S})$
          \\
          \midrule[0.875pt]
          $\emptyset$ &
          $\sfrac{13}{16}$ &
          $\sfrac{1}{4}$ &
          $-\sfrac{9}{16}$ &
          $\sfrac{0!(3-0-1)!}{3!}=\sfrac{1}{3}$ &
          $-\sfrac{9}{48}$
          \\
          $\{1\}$ &
          $\sfrac{1}{2}$ &
          $\sfrac{1}{2}$ &
          $0$ &
          $\sfrac{1!(3-1-1)!}{3!}=\sfrac{1}{6}$ &
          $0$
          \\
          $\{2\}$ &
          $\sfrac{5}{8}$ &
          $\sfrac{1}{4}$ &
          $-\sfrac{3}{8}$ &
          $\sfrac{1!(3-1-1)!}{3!}=\sfrac{1}{6}$ &
          $-\sfrac{3}{48}$
          \\
          $\{1,2\}$ &
          $\sfrac{1}{2}$ &
          $\sfrac{1}{2}$ &
          $0$ &
          $\sfrac{2!(3-2-1)!}{3!}=\sfrac{1}{3}$ &
          $0$
          \\
          \midrule[0.75pt]
          \multicolumn{5}{r}{SHAP score for feature 3 \hfill
            $\sv(3)~~=$} &
          $-0.25$ \\ 
          \bottomrule[1pt]
        \end{tabular}
      \end{center}
    \end{subfigure}
    
    \smallskip

    \caption{Computation of SHAP scores for the example DT and
      instance $((1,1,\sfrac{1}{2}),\alpha)$. For each feature $i$,
      the sets to consider are all the sets that do not include the
      feature. The expected values are those shown in~\cref{fig:cfs01}.
    }
    \label{fig:svs01}
\end{figure*}

\begin{exmpl}
  Using the expected values from~\cref{fig:cfs01}, \cref{fig:svs01}
  summarizes the computation of the (exact) SHAP scores. As can be
  concluded, the features that are irrelevant for the prediction, have
  non-zero SHAP scores, and the only feature that is relevant for the
  prediction has a SHAP score of 0.
  For a human decision-maker, it would be in error to deem features 2
  and 3 and having any influence in the prediction.
\end{exmpl}

A wealth of similarly unsatisfactory results are analyzed in several
recent
works~\cite{hms-corr23a,hms-corr23c,hms-corr23d,msh-corr23,msh-cacm24,hms-ijar24}.
Motivated by these results, several other works have
investigated both alternatives to SHAP
scores~\cite{ignatiev-corr23a,izza-corr23,ignatiev-corr23b,izza-aaai24},
or to their definition~\cite{lhms-corr24,lhams-corr24}.

\subsection{The Misconception of (Global) Adversarial Robustness}


For more than a decade~\cite{szegedy-iclr14,szegedy-iclr15},
researchers have witnessed that NNs are \emph{brittle}, i.e.\ small
changes on the input of an NN can results in unexpected changes in
prediction. In the case of images, the changes in prediction can defy
what human decision-makers would expect to see predicted. Such
examples are referred to as \emph{adversarial examples}.
Motivated by these observations, a line of research has been to
enforce and validate the (adversarial) robustness of ML models, thus
ensuring the non-existence of adversarial examples.
At present, the vast number of robustness methods is regularly
assessed in a dedicated competition~\cite{johnson-sttt23}.
Unfortunately, recent work~\cite{ims-corr23} demonstrated that
although local robustness can be correctly decided, global robustness
cannot, and so adversarial robustness can only be validated against a 
pre-determined number of test cases.
Despite these negative result, it has also been shown that adversarial
robustness is a stepping stone to deliver distance-restricted
explainability.

\section{A Glimpse of the Future} \label{sec:glimpse}

\subsection{Explaining Large-Scale ML Models}


Since the inception of logic-based
XAI~\cite{darwiche-ijcai18,inms-aaai19}, a major limitation has been
the issue of scalability, given the computational complexity of
computing one abductive explanation for the most widely used ML
models, e.g.\ NNs.
This limitation was apparent given early results on
NNs~\cite{inms-aaai19}, and acknowledged in overviews of the
field~\cite{ms-rw22}.

However, recent work proposed the novel concept of distance-restricted
(abductive and contrastive) explanations~\cite{hms-corr23b}.
distance-restricted explanations trade off the global validity of
explanations by a localized definition of explanation that can be
computed using off-the-shelve tools for deciding adversarial
robustness.%
\footnote{
The relationship between abductive explanations and adversarial
examples was first studied in earlier work~\cite{inms-nips19}.}
Furthermore, a stream of recent
works~\cite{barrett-corr22,hms-corr23b,barrett-nips23,ihmpims-corr24}
have significantly improved the scalability of computing
(distance-restricted) explanations in the case of complex NN models.

Perhaps more importantly, distance-restricted explainability
demonstrates that computing explanations is computationally
s difficult as deciding adversarial robustness, modulo the number of
features of the ML model.

\subsection{From Feature Selection to Feature Attribution}



As illustrated earlier, SHAP~\cite{lundberg-nips17} is based on a
definition of Shapley values~\cite{shapley-ctg53} that will produce
unsatisfactory results. A recent sequence of reports and
papers~\cite{hms-corr23a,hms-corr23c,hms-corr23d,msh-corr23,msh-cacm24,hms-ijar24}
has provided several examples of classifiers for which such
unsatisfactory results are obtained. Moreover, this paper illustrates
that the same issues occur with regression models
(see~\cref{ssec:svflaws}).

There is recent
work~\cite{ignatiev-corr23a,izza-corr23,ignatiev-corr23b,izza-aaai24,lhms-corr24,lhams-corr24}
that addresses the issues with Shapley values in XAI, and their uses
in SHAP.
Most of the proposed solutions show a tight connection between the
rigorous computation of explanation by feature selection, i.e.\ AXps
and CXps, and feature importance
scores~\cite{lhms-corr24,lhams-corr24}, which capture rigorously
defined measures of feature importance.
Furthermore, some recent
results~\cite{ignatiev-corr23a,izza-corr23,ignatiev-corr23b,izza-aaai24,lhams-corr24}
propose the use of AXps as a mechanism for computing measures of
feature importance, which represent alternatives to the use of Shapley
values in XAI.
The near future is expected to see further work along the same lines
of research.

\subsection{Certified Explainability}

%

One of the halmarks of automated reasoning has been the certification
of a reasoners'
answers~\cite{heule-cacm17,cfmssk-jar19,nordstrom-cade23}.
Recent work proposed the certification of computed explanations in
XAI~\cite{hms-tap23}.
However, this initial work~\cite{hms-tap23} focused on monotonic
classifiers. Future research will devise solutions for the
vertification of computed explanations, but also of tools that answer
other explainability queries.

\subsection{Additional Topics}


The field of logic-based XAI is bristling with new directions of
research.
One example is to stufy the conditions for the use of surrogate models
for computing the explanations of complex ML models. Although there
exists initial work~\cite{mazure-cikm21}, it is unclear how rigorous
the computed explanations are with respect to the starting complex
model.
Additional examples include, for example, the measure of rigorous
feature attribution~\cite{lhams-corr24} in the case of
distance-restricted explanations~\cite{bmpms-corr23}.

\section{Conclusions} \label{sec:conc}

This paper overviews the emergence of logic-based XAI, covering its
progress and main results, but also discussing some of its remaining
limitations. Furthermore, the paper also lists several ongoing
promising research directions, which address some of the remaining
challenges of logic-based XAI.
Finally, the paper also highlights how logic-based XAI has served to
uncover several misconceptions of non-rigorous XAI.

\jnoteF{Just wrap-up. Repeat some of the key take-home messages.}

%

\subsubsection*{Acknowledgments.}
%
This work was funded in part by starting funds provided by ICREA.
Over the years, this work received the input of several colleagues,
that include 
A.~Ignatiev, N.~Narodytska, M.~Cooper, Y.~Izza, X.~Huang, 
O.\ L\'{e}toff\'{e}, A.\ Hurault, R.\ Passos, A. Morgado, J.\ Planes,
R. B\'{e}jar, 
C.~Menc\'{\i}a 
and N.~Asher, among others.
Finally, the author also acknowledges the 
incentive provided by the ERC in not funding this research.


\newtoggle{mkbbl}

\settoggle{mkbbl}{false}

\iftoggle{mkbbl}{
  \bibliographystyle{splncs04}
  \bibliography{refs}
}{

}
%
\let\addcontentsline\oldaddcontentsline

\end{document}